%% file: Neural-Relightable.tex
\documentclass[acmtog,nonacm]{acmart}

\usepackage{booktabs} 

\usepackage{float}
\usepackage[ruled]{algorithm2e} 

\SetAlFnt{\small}
\SetAlCapFnt{\small}
\SetAlCapNameFnt{\small}
\SetAlCapHSkip{0pt}
\usepackage{soul}

\acmJournal{TOG}

\newcommand{\wz}[1]{{\color{red}{[Wenzheng: #1]}}}
\newcommand{\wyj}[1]{{\color{blue}{[Youjia: #1]}}}

\input{section/notation}

\begin{document}
\title{NARRATE: A Normal Assisted Free-View Portrait Stylizer}

\author{Youjia Wang}
\orcid{0000-0002-0517-3475}
\affiliation{%
 \institution{ShanghaiTech University}
 \country{China}}
\email{wangyj2@shanghaitech.edu.cn}
\author{Teng Xu}
\affiliation{%
 \institution{ShanghaiTech University}
 \country{China}
}
\email{xuteng@shanghaitech.edu.cn}
\author{Yiwen Wu}
\affiliation{%
\institution{ShanghaiTech University}
\country{China}
}
\email{wuyw1@shanghaitech.edu.cn}
\author{Minzhang Li}
\affiliation{%
 \institution{ShanghaiTech University}
 \country{China}
}
\email{limzh@shanghaitech.edu.cn}
\author{Wenzheng Chen}
\affiliation{%
 \institution{University of Toronto}
 \country{Canada}
 }
\email{wenzheng@cs.toronto.edu}
\author{Lan Xu}
\affiliation{%
 \institution{ShanghaiTech University}
 \country{China}
}
\email{xulan1@shanghaitech.edu.cn}
\author{Jingyi Yu}
\affiliation{%
 \institution{ShanghaiTech University}
 \country{China}
}
\email{yujingyi@shanghaitech.edu.cn}

\input{section/abs}

%
%

	
	%
	%


\begin{teaserfigure}
  \includegraphics[width=\textwidth]{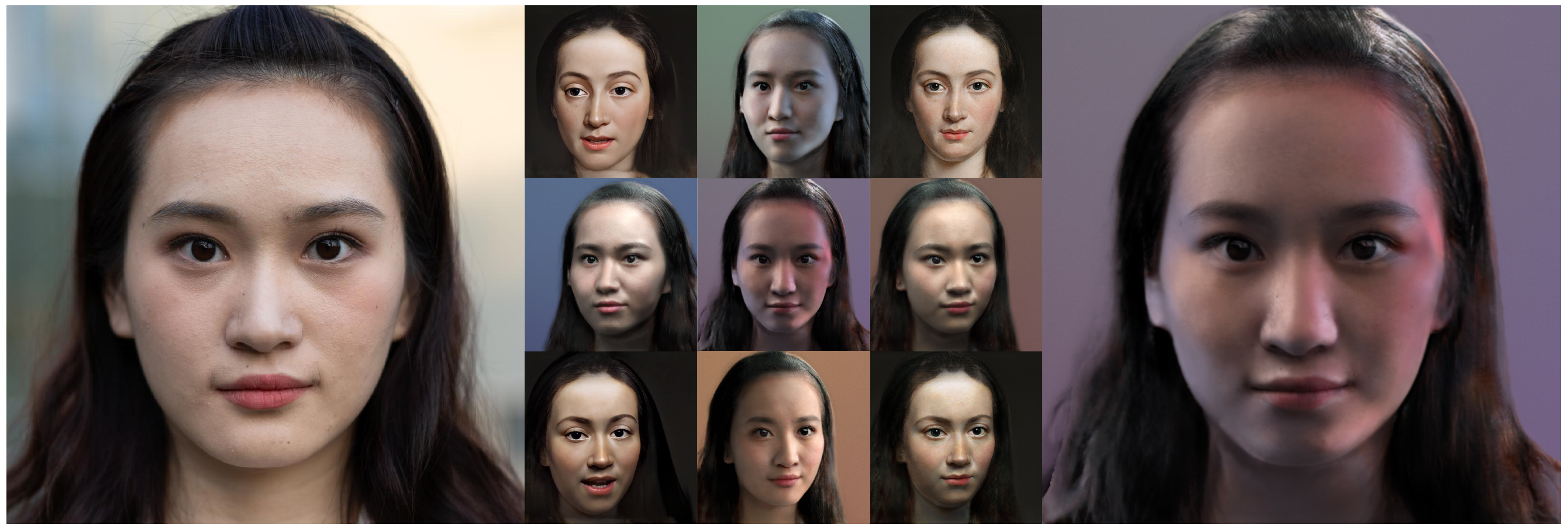}
  \caption{We propose {\ourmodel}, a novel portrait stylization pipeline that supports editing portrait lighting and perspective  in a photorealistic manner.  Taking a single high resolution portrait as input, it conducts pose change, light change,
facial animation, and style transfer, either separately or in combination, at a photographic quality. }
  \label{fig:teaser}
\end{teaserfigure}

\maketitle

\input{section/Introduction}

\input{section/Related_work}
\input{section/Method}

\input{section/Experiments}

\section{Conclusion}

In this paper, we present {\ourmodel}, a novel portrait stylization pipeline that enables high-quality perspective editing and relighting.
{\ourmodel} leverages a hybrid neural-physical face model to generate photorealistic images under novel perspectives, as well as consistent input normal maps in relighting to create coherent new illumination effects.
It can further couple with various facial applications, boosting them in free-view and relightable manner. 

We experimentally demonstrate that {\ourmodel} produces more plausible and stable view editing and relighting results, outperforming existing state-of-the-art methods by a large margin.
To the best of our knowledge, {\ourmodel} is the first approach that supports perspective and lighting adjustment at photographic quality. We believe it opens a new door for portrait stylization and has great potential to facilitate various AR/VR applications like virtual cinematography, 3D video conferencing, and post-production.

\bibliographystyle{ACM-Reference-Format}
\bibliography{sample-bibliography}

\end{document}

%% file: section/notation.tex
\newcommand{\ourmodel}{NARRATE}
\newcommand{\stylesdf}{StyleSDF}
\newcommand{\fomm}{FOMM}

\newcommand{\image}{\mathbf{I}}
\newcommand{\imageNote}[1]{\image^{#1}}
\newcommand{\imageNoteUp}[2]{\image^{#1}_{#2}}
\newcommand{\view}{\mathbf{v}}
\newcommand{\viewarb}{\tilde{\view}}

\newcommand{\normal}{\mathbf{N}}
\newcommand{\normalNote}[1]{\normal_{#1}}
\newcommand{\normallowqual}{\normalNote{l}}
\newcommand{\normalhighqual}{\normalNote{h}}

\newcommand{\neural}{\mathbf{n}}
\newcommand{\mesh}{\mathbf{m}}

\newcommand{\latentz}{\mathbf{z}}

\newcommand{\generator}{\mathcal{G}}
\newcommand{\generatorparam}{\mathbf{\theta}}

\newcommand{\meshNote}[1]{\mesh_{#1}}
\newcommand{\meshlowqual}{\meshNote{l}}
\newcommand{\meshhighqual}{\meshNote{h}}

\newcommand{\loss}{\mathcal{L}}
\newcommand{\lossNote}[1]{\loss_{#1}}
\newcommand{\lossMSE}{\lossNote{MSE}}
\newcommand{\lossPer}{\lossNote{Per}}

\newcommand{\lweight}{\lambda}
\newcommand{\lweightNote}[1]{\lweight_{#1}}
\newcommand{\lweightMSE}{\lweightNote{MSE}}
\newcommand{\lweightPer}{\lweightNote{Per}}

\newcommand{\network}{\mathcal{F}}
\newcommand{\networkNormal}{\network_N}
\newcommand{\networkAlbedo}{\network_A}
\newcommand{\networkRelighting}{\network_R}

\newcommand{\imageAlbedo}{\image_A}
\newcommand{\imageRelit}{\image_R}

\newcommand{\imageAlbedoUp}[1]{\imageAlbedo^{#1}}
\newcommand{\imageRelitUp}[1]{\imageRelit^{#1}}
\newcommand{\normalUp}[1]{\normal^{#1}}

\newcommand{\light}{L}
\newcommand{\lightEnv}{\light_e}

\newcommand{\normalhighqualUp}[1]{\normalhighqual^{#1}}

\newcommand{\videoTar}{D}
\newcommand{\framet}{t}
\newcommand{\videoTarNote}[1]{\videoTar_{#1}}
\newcommand{\motionField}{\mathbf{T}}
\newcommand{\motionFieldNote}[2]{\motionField_{{#1}\leftarrow{#2}}}
\newcommand{\occlusionField}{\mathbf{O}}
\newcommand{\occlusionFieldNote}[2]{\occlusionField_{{#1}\leftarrow{#2}}}
\newcommand{\networkMotionGenerator}{\mathbf{G}}

\newcommand{\networkStyleOil}{\mathbf{M}}
\newcommand{\style}{\mathbf{S}}
\newcommand{\shading}{S}

%% file: section/abs.tex
\begin{abstract}
	
	
	 Taking a compelling portrait photo relies on deliberately designed lighting styles and shooting angles, requiring advanced skills beyond the ability of casual users. 
	%
	Building a portrait stylization tool that automates lighting and perspective editing 
	is beneficial to photographers and artists, enabling numerous applications from photography/cinematography to AR/VR. 
	%
	This task remains challenging as \emph{persuasive lighting adjustment}, \emph{geometrically correct perspective changes}, and \emph{high photorealism maintenance} need to be achieved at the same time.
	Naively bridging existing relighting and view synthesis methods 
	produces inaccurate and unstable 
	results.

	In this work, we propose {\ourmodel}, a novel pipeline that enables simultaneously editing portrait lighting and perspective in a photorealistic manner. 
	As a hybrid neural-physical face model, {\ourmodel} leverages complementary benefits of geometry-aware generative approaches~\cite{orel2021styleSDF} and normal-assisted physical face models.
	In a nutshell, {\ourmodel} 
	first inverts the input portrait to a coarse geometry and employs neural rendering to generate images resembling the input, as well as producing convincing pose changes.
	However, inversion step introduces mismatch, 
	bringing low-quality images which fail to keep facial details.
	As such, we further estimate portrait normal to enhance the coarse geometry, creating a high-fidelity physical face model to render more detailed images.
	In particular, we fuse the neural and physical renderings to compensate for the imperfect inversion, resulting in both realistic and view-consistent novel perspective images.
	
	In relighting stage, previous works~\cite{sun2019single,pandey2021total} focus on single view portrait relighting but ignoring consistency between different perspectives as well, leading unstable and inconsistent lighting effects for view changes.  
	We extend~\cite{pandey2021total} to fix this problem by unifying its multi-view input normal maps  
	with the physical face model. 
	{\ourmodel} conducts relighting with consistent normal maps, imposing cross-view constraints and exhibiting stable and coherent illumination effects.

	We experimentally demonstrate that {\ourmodel} achieves 
	more photorealistic, reliable results over prior works. 
	We further bridge {\ourmodel}  with animation and style transfer tools, supporting pose change, light change, facial animation, and style transfer, either separately or in combination, all at a photographic quality. We showcase vivid free-view facial animations as well as 3D-aware relightable stylization, which help facilitate various AR/VR applications like virtual cinematography, 3D video conferencing, and post-production.
\end{abstract}

%% file: section/Introduction.tex
\section{Introduction}
Even in today's digital era, portrait photography remains inarguably the most popular genre. 
For any successful portrait, be it an actor still, a beauty shot, or a character study, \emph{lighting styles}, coupled with \emph{shooting angles}, play the key role to emphasize the subject's feature as well as to express artistic or dramatic intent. On lighting, popular loop lighting defines facial features with soft shadows to illustrate harmony whereas a half lit and half in shadow face by split lighting reflects the subject's inner struggle. Rembrandt lighting combines a key light and a fill light (e.g., a reflector) to cast a distinctive triangle of light on one cheek, baking a moody atmosphere. On shooting angle, front view (full face) shots highlight complete facial details to convey an assertive attitude whereas side view (profile) shots emphasize on the physical form-factor such as facial contours to exude character. Three-quarter views, popularized by self-portraits, reveal both with a personal touch. 



Despite richness in styles, a portrait photograph, once taken, is fixed in its final presentation.
Editing its lighting and perspective to produce an equally compelling alternative, if possible at all, requires tremendous manual processing even for the professionals. Overall, restylizing the photograph not only requires using a disarray of advanced retouching tools but can be extremely time-consuming.
As such, an automated portrait stylization pipeline is highly valuable for photographers and visual artists as well as for daily users. Yet the challenges are multi-fold, including \textbf{geometrically correct perspective changes}, \textbf{persuasive lighting adjustment}, and \textbf{high photorealism maintenance}.  

Face pose changes have long followed the warping pipeline that largely relies on accurate 3D facial geometry estimations. Exemplary techniques such as 3D morphable models (3DMM)~\cite{blanz1999morphable} can recover large scale geometry but tend to produce overly smooth surfaces that lack details. Further, 3DMM cannot model geometry beyond face such as hair, clothing, glasses, and even ears and consequently the morphing results exhibit strong visual artifacts in respective regions. Recent implicit solutions, popularized by generative models~\cite{gan2014}, can generate unseen portraits with convincing pose variations~\cite{karras2019style,shen2020interpreting}. However, as a generator, they do not readily support restylizing a given portrait. Techniques such as inversion~\cite{zhu2016generative} for fitting a latent code to the known image can lead to excessive blurs or distortions ~\cite{tov2021designing}.

Adjusting lighting, on the other hand, requires reliable estimations of a portrait's original illumination, facial reflectance, and facial normal. 
Existing portrait relighting techniques have largely focused on fixed viewpoint retouching ~\cite{sun2019single,pandey2021total}. When applied to new facial poses, e.g., rigged using 3DMM, they produce strong visual artifacts with even slight illumination and shape inaccuracy and continuous pose changes generate strong flickering.

In this paper, we present {\ourmodel} (\textbf{N}ormal \textbf{A}ssisted \textbf{P}ortrait \textbf{St}yliz\textbf{er}), a novel portrait stylization pipeline 
that takes a single high resolution portrait as input and conducts pose change, light change, facial animation, and style transfer, either separately or in combination, at a photographic quality. 
At the core of {\ourmodel} is a hybrid neural-physical face model that leverages complementary benefits of geometry-aware generative approaches ~\cite{orel2021styleSDF} and normal-assisted physical face models.
In a nutshell, {\ourmodel} first conducts GAN inversion~\cite{roich2021pivotal} for finding a generated portrait that closely resembles the input. 
In 3D-aware generation, the inverted portrait latent initiates a coarse geometry in order to produce convincing pose changes, as well as employs neural rendering to synthesize complete portrait effects including hair, ear, and clothing.
However, GAN inversion brings mismatch between the input and fitted images, resulting in low-quality, less detailed synthesis. Thus, we further adopt high fidelity normal estimation techniques to infer portrait albedo and normal information~\cite{pandey2021total} to enhance the coarse geometry to a high quality physical face model, synthesizing images full of realistic facial details with graphics rendering engines.
In particular, we fuse the two renderings to composite more realistic and consistent novel perspective images.

For relighting, we extend the fixed viewpoint framework~\cite{pandey2021total} to support consistent relighting over continuous viewpoint changes. 
Instead of processing individual poses, 
We show how to conduct neural relighting on all views at once 
by imposing consistent normal and albedo maps from the same physical face model. 
Our technique significantly suppresses flickering than per-frame based baseline where a user can specify exquisite lighting effects (e.g, Rembrandt) at any new poses to produce photographic portraits or videos.

{\ourmodel} also supports dynamic animations and non-photorealistic stylization. Combined with either implicit or explicit animators~\cite{fomm2019}, {\ourmodel} produces relightable free-viewpoint videos of virtual human talents such as opera singing, potentially useful for virtual cinematography and post-production. 
We also bridge  {\ourmodel} with artistic style transfer tools~\cite{wang2018pix2pixHD, praun2001real}, creating 3D-aware styled images like oil painting and hatching drawing, bringing famous portrait paintings to life.

We summarize our contributions as follows:
\begin{itemize}
	\item We propose {\ourmodel}, a novel portrait stylization pipeline that supports editing perspective and lighting in a photorealistic manner.
	
	\item Our hybrid neural-physical face model takes advantage of geometry-aware generative methods and normal-assisted physical models, fusing neural and graphics renderings to create plausible and stable free-view relighting effects.
	
	\item {\ourmodel} can further combine with animation and style transfer tools to synthesize delicate styled facial animation, demonstrating its potential in future AR/VR applications.
\end{itemize}

%% file: section/Related_work.tex
\section{Related Work}
Our work focuses on post-capture photographic portrait adjustment and retargeting. While previous approaches have adopted different strategies to handle pose, lighting, style, and animations, we propose a unified solution via hybrid neural physical face modeling.  
\paragraph{Perspective Adjustment.}
Post-capture perspective adjustment to portrait has drawn broad interest in computer graphics and vision, and latest virtual reality. Images are essentially 2D projections of 3D geometry and therefore previous efforts have focused on inferring and then rigging 3D faces from the portraits. Since the problem is inherently ill-posed, earlier approaches attempt to fit 3D template models (3DMM)~\cite{blanz1999morphable} from landmarks via optimization schemes~\cite{blanz2004exchanging,cao2014displaced,cao2015real}. More recent methods ~\cite{kazemi2014one,zhu2017face,tewari2017mofa,tewari2018self,richardson20163d,richardson2017learning,yamaguchi2018high} adopt a data-driven strategy and produce more faithful reconstructions on both shape and appearance. By far, existing parametric template models~\cite{blanz1999morphable,li2017learning,feng2020learning} cannot yet preserve fine-grained facial details. They also only model the facial regions while ignoring other body parts such as hair styles or even ears. Pose adjustments on high quality portrait often lead to uncanny results where faces look plastic due to loss of details.

The emerging implicit solutions, popularized by generative models~\cite{gan2014,karras2019style,Karras2019stylegan2,Karras2021stylegan3}, have shown huge potential for free-view high-quality portrait synthesis. Although, as a generation task, they were not specifically designed for perspective retargeting, it is possible to tweak these solutions via neural inversion to support free-view rendering \cite{karras2019style,tewari2020stylerig,shen2020interpreting,shen2021closedform}. Yet, despite significant advances, neural inversion still loses high frequency and earlier generative methods produce aliased and inconsistent view changes.
Followup geometry-aware generative models~\cite{orel2021styleSDF,chanmonteiro2020pigan,ed3d,zhou2021CIPS3D,gu2021stylenerf} combine 2D generation with explicit 3D geometry constraints and have greatly improved view consistency. However, their estimated geometry is still too smooth and lacks the precision for conducting other stylization tasks such as relighting. Most recently,  ~\cite{tan2022volux} imposes additional lighting constraints to simultaneously achieve relighting and perspective changes. Yet, the final quality still falls short of the photographic level due to volumetric neural rendering which also causes cross-view flickering . 
%

\paragraph{Portrait Relighting.}

A separate research line on portrait stylization is one-shot relighting. Pioneered by the LightStage ~\cite{debevec2000acquiring}, approaches over the past decade have focused on capturing and modeling the 4D reflectance field. Using a spherical lighting rig, various version of the LightStage manage to capture one-light-at-a-time or OLAT reflectance field \cite{pandey2021total, zhang2021neural, guo2019relightables, meka2019deep} for a human face and even their full body ~\cite{meka2020deep}. 
Latest advances on deep learning have subsequently enabled efficient modeling of the OLAT data to address various aspects of the relighting problem ~\cite{shu2017portrait,meka2019deep,meka2020deep,guo2019relightables,zhou2019deep,sun2019single,nestmeyer2020learning,wang2020single,pandey2021total}, ranging from modeling lighting using spherical harmonics representations ~\cite{zhou2019deep,liu2021relighting,sengupta2018sfsnet} to more sophisticated environment lighting \cite{sun2019single}. To overcome the limitation of low frequency lighting, \cite{nestmeyer2020learning} explicitly models shadows and specularity by directional light source whereas \cite{wang2020single} employs richer synthetic data to account for non-Lambertian effects. The seminal work of TotalRelighting \cite{pandey2021total} significantly improves prior learning based techniques by employing specialized network designs coupled with robust matting. Their technique can produce photographic quality results under a broad range of virtual lighting. Yet, TotalRelighting is still restricted to fixed viewpoint. Brute-force application to pose adjusted portrait can lead to incorrect normal estimation and subsequently visual artifacts.

\paragraph{Animation.}
An exciting new area of portrait stylization is to animate the face via a driving video. The solution has largely followed facial motion transfers or reenactment and has been divided into subject-dependent and subject-agnostic models, both can be conducted either in 3D or 2D. 3D approaches~\cite{fried2019text,geng2018warp,nagano2018pagan,olszewski2017realistic} are more robust to pose changes and potentially amenable to relighting. However, since reliable reconstructions can only be conducted on near diffuse regions with simple geometry, they often miss teeth, ear, hair etc, causing the reenacted performance unconvincing. A variety of 2D generative models~\cite{fomm2019,zakharov2019few,zakharov2020fast,wang2019few} can now produce highly compelling facial animations without using any 3D shape prior. The lack of 3D, however, limits them from perspective changes, let alone relighting which we aim to resolve. The latest Neural Talk Head~\cite{wang2021one} demonstrates that even slight support of 3D in the general framework can lead to big improvement. By using canonical 3D keypoints, they are able to produce free-view video synthesis from a single face image. Their quality may be sufficient for teleconferencing but falls short of production-level portrait photography. 

\paragraph{Artistic Stylization.} Artistic stylization has a long history in rendering, ranging from early statistics model ~\cite{tang2003face,shih2014style} to the deep learning model ~\cite{gatys2015neural,johnson2016perceptual,liao2017visual,selim2016painting}. Latest solutions essentially model the problem as image-to-image translation ~\cite{CycleGAN2017,pix2pix2017,futschik2019real}. As they unanimously assume fixed viewpoint input \cite{chen2018stereoscopic,d2017coherent,Jamriska19-SIG}, combining these solutions with free-view rendering require high consistent across viewpoints. Recent techniques ~\cite{fivser2017example,han2021exemplar} attempt to impose 3D constraints either based on a depth map or an explicit 3D model, yet both lacked precision to produce faithful perspective changes while maintaining fine details. We instead propose a hybrid neural physical model that can be adopted to simultaneously tackle perspective change, relighting, animation, or artistic stylization.

%% file: section/Method.tex
\begin{figure*}[t]
	\begin{center}
		\includegraphics[width=0.93\linewidth]{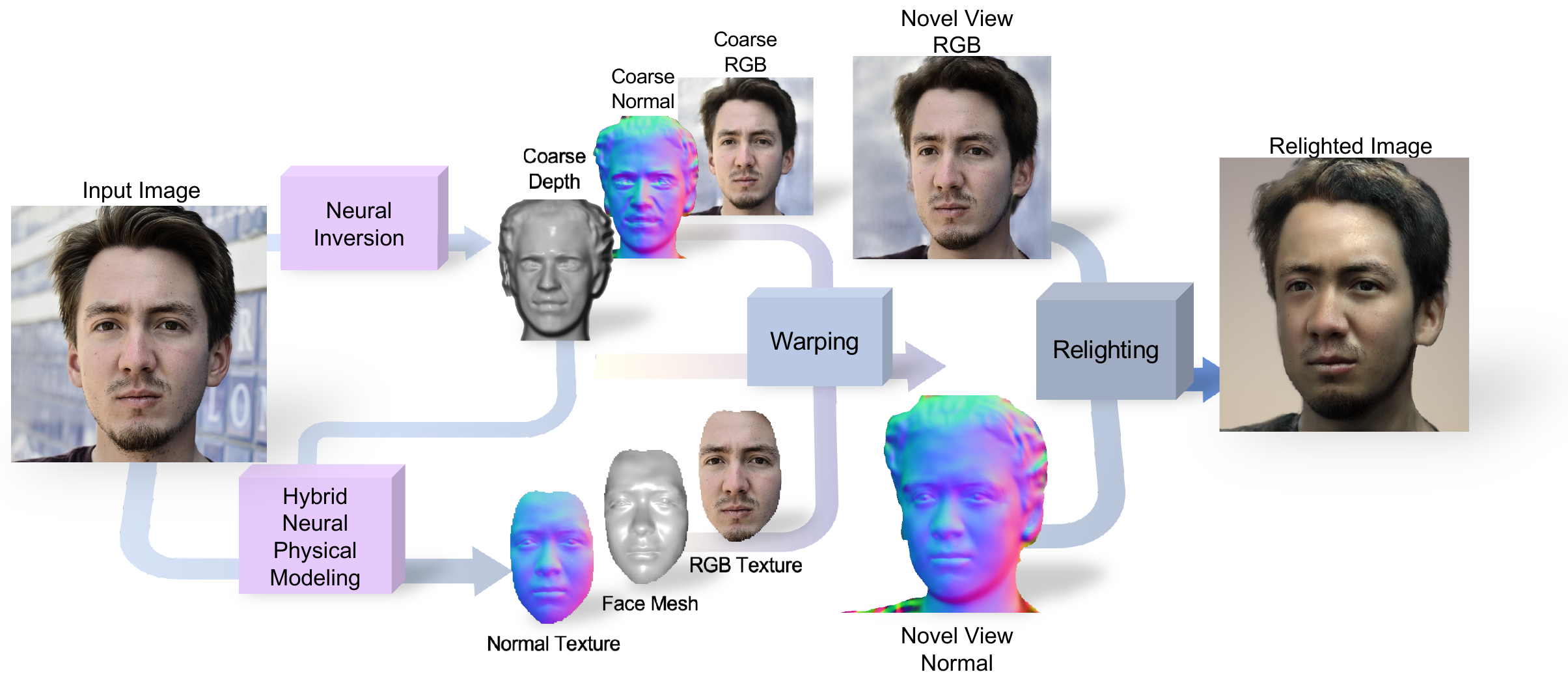}
	\end{center}
	\vspace{-0.5cm}
	\caption{
	Taking a portrait as input (left), {\ourmodel} first performs neural inversion(Sec.~\ref{sec:invertion}) to produce a coarse geometry and less-detailed neural rendered images(middle left, top).
		It further creates a high-fidelity physical face model(Sec.~\ref{sec:normal}) to render more detailed images(middle left, bottom).
		{\ourmodel} warps the two renderings(Sec.~\ref{sec:fusion}), generating realistic and consistent novel view images as well as accurate normal maps (middle right), which are readily used to create plausible relighting effects (right).}
	\label{fig:overview}
	\vspace{-0.3cm}
\end{figure*}

\section{Overview}
\label{sec:overview}

In this section, we describe {\ourmodel} pipeline and applications.
As shown in Fig.~\ref{fig:overview}, given a portrait image, {\ourmodel} allows separate modifications to its perspective and lighting, enabling photorealistic and consistent editing effects at a photographic quality across different views.
{\ourmodel} can further integrate with animation and style transfer tools to create free-view animated/styled faces, 
benefiting AR/VR applications such as tele-presence as well as virtual cinematography.
We first describe photorealisitc free view synthesis and view-consistent relighting in Sec.~\ref{sec:perspective}. Then, we present free-view, relightable facial animation and style transfer applications in Sec.~\ref{sec:application}.

\section{Free-View Relight}\label{sec:perspective}





Take a portrait $\image$ as input, we set out to synthesize a novel view $\imageNote{\view}$ under a new camera perspective $\view$ while maintaining photorealism. 
{\ourmodel} adopts a hybrid neural-physical face model by leveraging a geometry-aware generative model and a normal-assisted physical face model.
As shown in Fig.~\ref{fig:overview} (left top), we first employ the powerful geometry aware generative models~\cite{orel2021styleSDF} and apply GAN inversion~\cite{roich2021pivotal} to reverse the input to the latent code. The generator then maps it to a coarse geometry $\meshlowqual$ and conducts neural rendering to synthesize a high resolution novel view image $\imageNoteUp{\view}{\neural}$. We describe detailed procedures in Sec.~\ref{sec:invertion}

We observe though that both inversion and neural rendering introduce artifacts in novel view synthesis, creating over-smoothed and aliasing results. As such, we further propose to enhance the result with normal-assisted physical meshes. 
Specifically, we employ a normal estimation network (Sec~\ref{sec:fusion}) to obtain a detailed normal map $\normal$ from the input $\image$. With the assistance of $\normal$, we improve the generated coarse geometry $\meshlowqual$ through Poisson normal integration~\cite{horn1986variational}, producing a high quality physical model $\meshhighqual$. We then texture it with the original image and 
render a new image $\imageNoteUp{\view}{\mesh}$ using the traditional pipeline~\cite{woo1999opengl}.
Finally, we fuse $\imageNoteUp{\view}{\neural}$ and $\imageNoteUp{\view}{\mesh}$ and with Poisson blending~\cite{perez2003poisson} and generate a photorealistic image with much reduced artifacts while maintaining details. We discuss the physical model generation and fusion processes in Sec.~\ref{sec:normal} and Sec.~\ref{sec:fusion}, respectively.



\subsection{Neural Inversion}
\label{sec:invertion}

Geometry-aware generative models can now produce images under explicit geometry constraints, achieving better view consistency in novel view synthesis. 
We build our pipeline on top of the  {\stylesdf}~\cite{orel2021styleSDF} framework, currently viewed as the state-of-the-art.
StyleSDF first maps a latent code $\latentz$ to a coarse Signed Distance Field (SDF),  where a low resolution image is generated via volume rendering~\cite{wang2021neus} with a specific camera view $\view$. 
Different from 2D generation architectures~\cite{karras2019style,Karras2019stylegan2,Karras2021stylegan3}, its rendering process helps regularize generation with explicit geometrical constraints. This allows us to adjust $\view$ to render different viewpoints towards the same shape, resulting in consistent multi-view images.
Neural rendering (generally 2D CNN layers) is further employed to upsample the low resolution image to a high resolution one.
We denote the complete generation as $\imageNoteUp{\view}{\neural}=\generator(\latentz, \view; \generatorparam)$ where $\generator$ is the generator and $\generatorparam$ is its parameters. The SDF can also be used to extract a coarse mesh $\meshlowqual$ as well as a low quality normal map $\normallowqual$ that will be used in later steps.

Conceptually, {\stylesdf} can be treated as a neural face model.
While {\stylesdf} is not designed for any subject, we perform GAN inversion to invert the input image $\image$ back to the latent $\latentz$. Once $\latentz$ is ready, we can utilize the generator to synthesize novel view images $\imageNoteUp{\view}{\neural}$ under any camera view $\view$. We choose to use the state of the art method PTI~\cite{roich2021pivotal} technique. While PTI is designed for StyleGAN 2D inversion, we extend it for {\stylesdf} 3D inversion.

Specifically, PTI contains two stages. We first fix the weights $\generatorparam$ of the generator $\generator$ and then optimize the latent code $\latentz$ as well as the camera view $\view$ such that the generated image would best match the input. 
Notice that brute-force optimization over $\latentz$ generally produces blurred results~\cite{tov2021designing}. Alternatively, we can fix $\latentz$ and $\view$ but optimize the generator weights $\generatorparam$ to enforce the generated image closer to the input. In either case, we refine the initial results using normal maps in Sec.~\ref{sec:normal}. In our implementation, we adopt the same loss and regulation as PTI where more details can be found in ~\cite{roich2021pivotal}.

\begin{figure}[ht]
    \begin{center}
	    \includegraphics[width=0.93\linewidth]{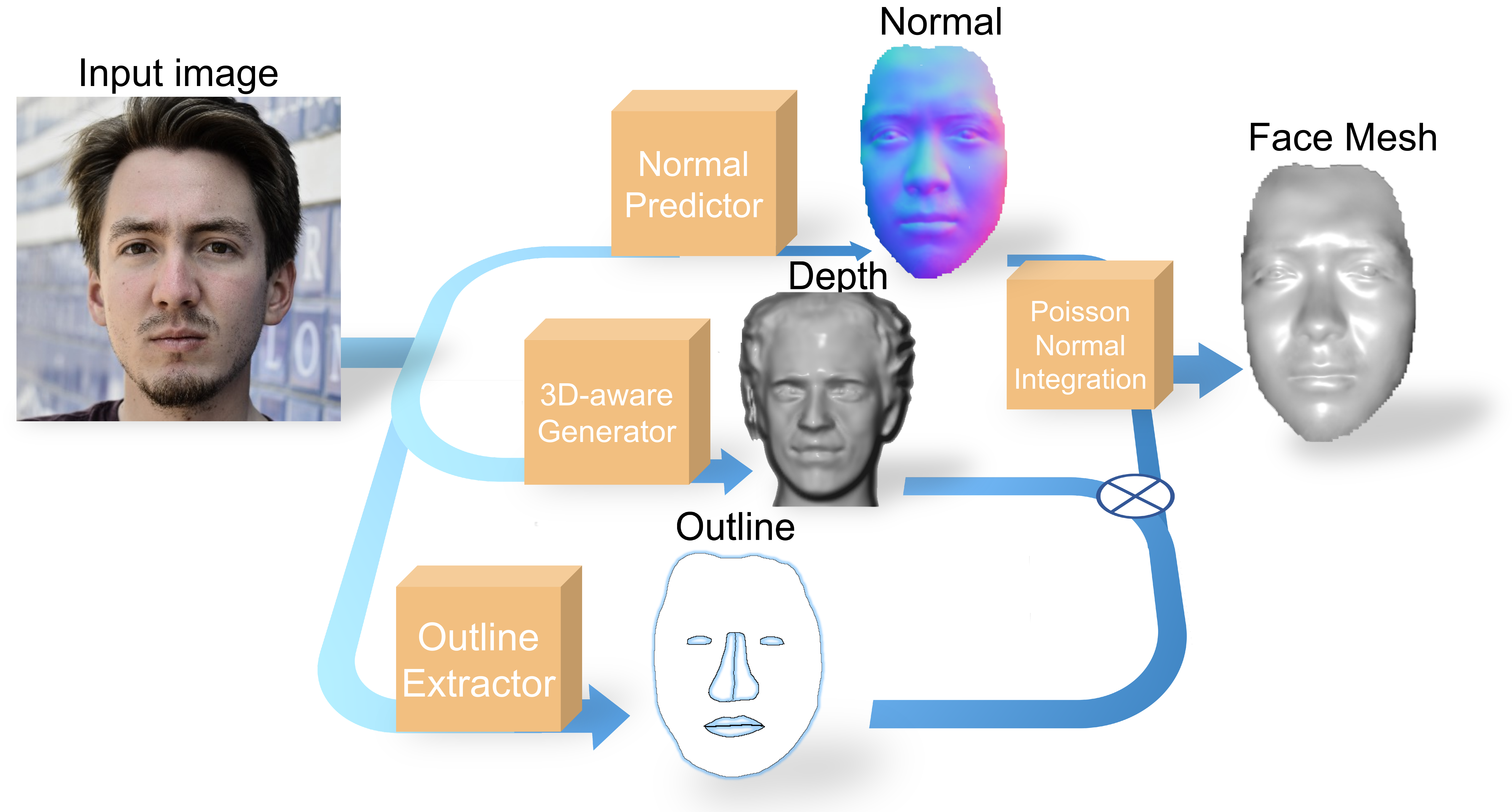}
    \end{center}
	\vspace{-0.5cm}
	\caption{
	{\ourmodel} creates its physical face model from high quality normal maps and depth priors. The former is inferred from a normal predictor while the latter comes from inverted coarse geometry. We perform Poisson normal integration for the outline region, generating a high-fidelity face mesh.}
	\label{fig:normal_estimation}
	\vspace{-0.3cm}
\end{figure}

\subsection{Hybrid Neural Physical Modeling}
\label{sec:normal}


Novel views synthesized from {\stylesdf} and PTI, although of a much higher quality than previous GAN-based results, still exhibit artifacts beyond acceptable as a photograph. The causes are two-fold: first, the image generated with PTI inversion still contains mismatch from the input portrait, especially for the regions with high frequency details. Second, precise view consistency only is only enforced on low resolution generated images whereas the upsampling process in neural renderings introduces additional aliasing and inconsistency, causing vibration under view changes when presented in a high resolution ~\cite{ed3d,chanmonteiro2020pigan}. 

The visual artifacts are magnified on portraits due to incredible visual acuity of human for distinguishing human faces. To mitigate the problem, we employ an auxiliary normal-assisted physical face model. 
Recall that {\stylesdf} generates not only 2D images but also the 3D geometry of the face, although it is coarse by nature. We thus aim to create a high quality face mesh, 
 textured with the original image, to enhance the facial details as well as reduce view inconsistency. We observe the coarse mesh $\meshlowqual$ extracted from {\stylesdf} is too noisy to be directly used a proxy. We therefore propose to enhance the geometric quality by exploiting extra normal information.

To do so, we set out to infer a high quality normal map $\normalhighqual$ from $\image$ via a normal predictor(Sec.~\ref{sec:fusion}), with which we then construct a high quality face mesh $\meshhighqual$ from $\meshlowqual$ with the assistance of $\normalhighqual$ by Poisson integration. Note that $\normalhighqual$ alone is insufficient for creating a high quality mesh $\meshhighqual$: discontinuous regions (e.g., nose, mouth) exhibit ambiguity that can lead to the reconstruction of a flattened 3D face. We hence solve a constrained linear equations, where $\meshlowqual$ provide depth priors in discontinuities whereas $\normalhighqual$ sets out to preserve fine geometric details. In our implementation, we focus on the front face as it contains more details. Next, we re-project $\meshhighqual$ to the image with camera view $\view$ and use the original image as the texture. We take $\meshhighqual$ as our normal-assisted physical model that supports being
rendered at arbitrary view $\viewarb$ with full details obtained from the front view. Fig. ~\ref{fig:normal_estimation} shows how $\meshhighqual$ is reconstructed.

\begin{figure}[t]
    \begin{center}
	    \includegraphics[width=0.93\linewidth]{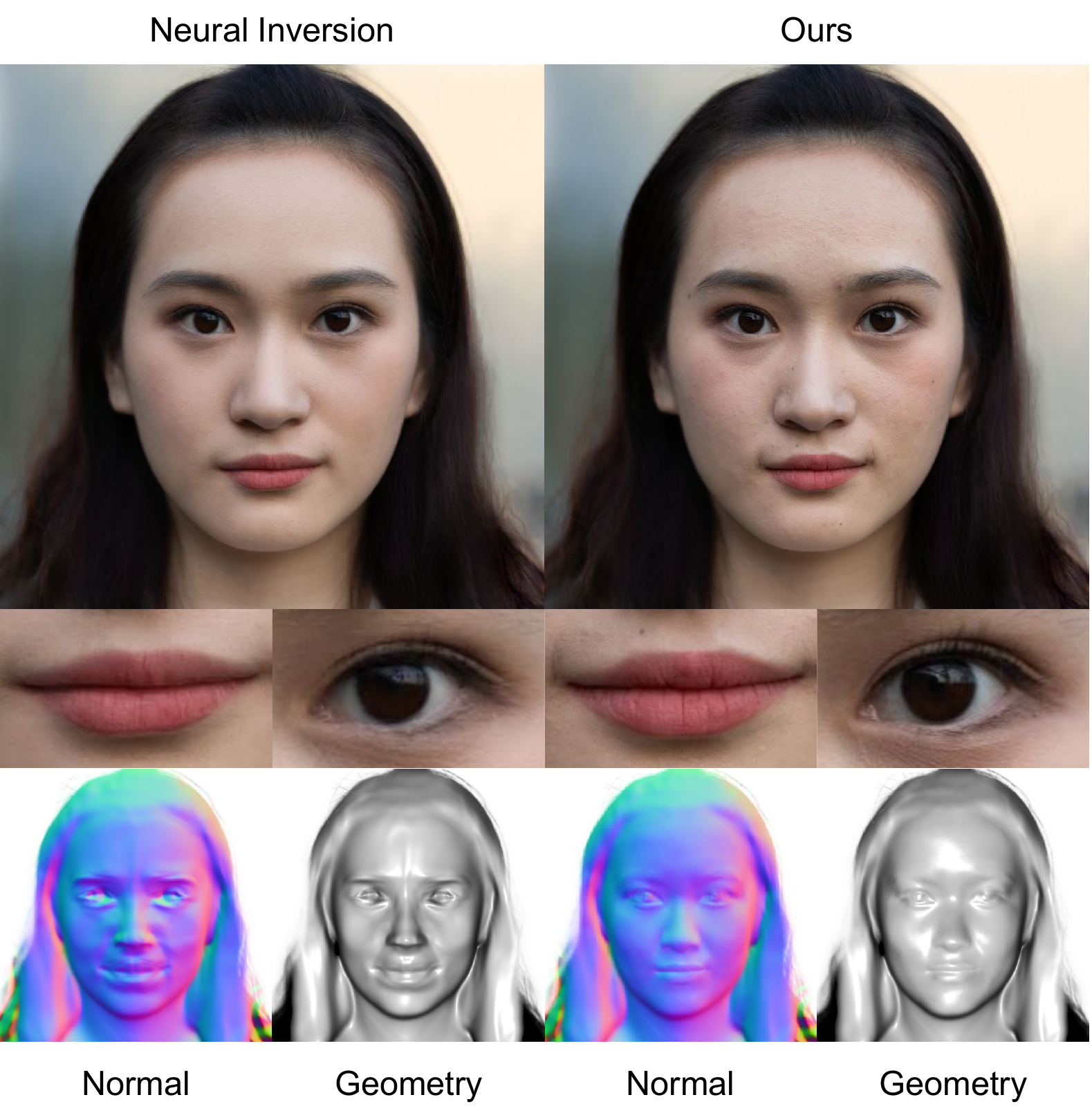}
    \end{center}
	\vspace{-0.5cm}
	\caption{
	Neural Inversion only generates images missing  facial details. {\ourmodel} compensates it by hybrid neural-physical model, achieving more detailed and realistic results
		In addition, it provides better surface normal maps, which helps in the subsequent relighting and other applications.}
	\label{fig:PTI_result}
	\vspace{-0.3cm}
\end{figure}

\subsection{Warping and Relighting}
\label{sec:fusion}



With renderings from both neural and physical models, we are able to create high quality, view-consistent images under arbitrary views. 
For a new camera view $\viewarb$, we synthesize a RGB image, as well as a normal map which is further used in subsequent relighting module and applications.
The final RGB image $\imageNote{\viewarb}$ is fused from the {\stylesdf} generation $\imageNoteUp{\viewarb}{\neural}$ and the mesh-rendered image $\imageNoteUp{\viewarb}{\mesh}$. $\imageNoteUp{\viewarb}{\neural}$ contains a complete portrait with all essential regions such as hairs, clothing, ears (and ear rings), and the even background that are critical for preserving photorealistic. At the same time, recall $\imageNoteUp{\viewarb}{\mesh}$ only presents the main face region but with fine details.

\begin{figure}[t]
	\begin{center}
		\includegraphics[width=0.93\linewidth]{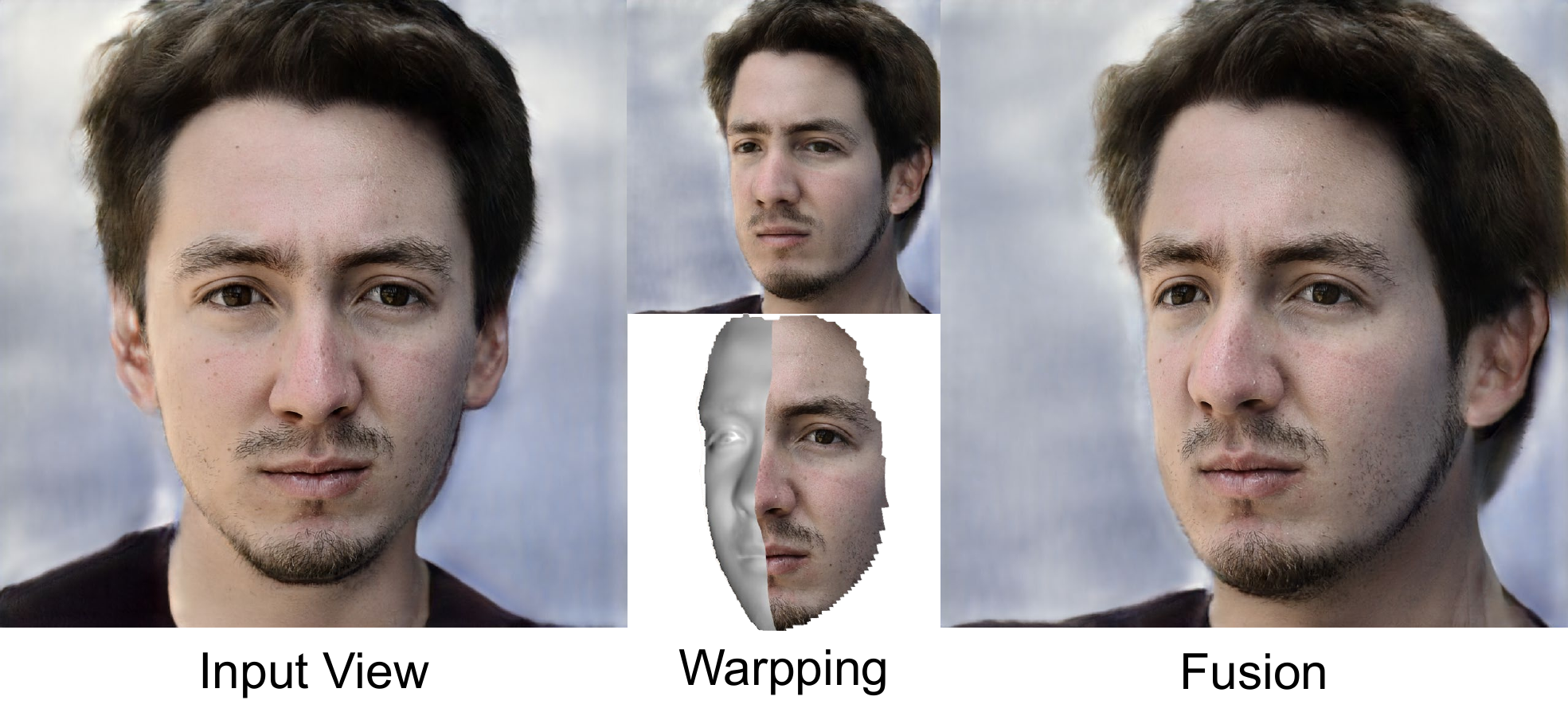}
	\end{center}
	\vspace{-0.5cm}
	\caption{
	Once the hybrid neural-physical model is ready, we can easily render novel view images from both neural part and physical part(middle). We then use Poisson image blending to fuse them and get the final image(right).}
	\label{fig:fusion_pipeline}
	\vspace{-0.3cm}
\end{figure}

We therefore create $\imageNote{\viewarb}$ by fusing the two by Poisson blending~\cite{perez2003poisson}. In our experiments, we observe the Poisson scheme marginally impacts photorealism but manages to maintain high consistency under view changes.
We further compute an companion normal map generated by applying the same fusion technique to two normal maps, the first from {\stylesdf} SDF whereas the other from rendering under camera pose changes with the normal map and mesh from $\meshhighqual$. Fig.~\ref{fig:PTI_result} compares our method with PTI where our technique significantly improves the visual fidelity in both detail preservation and noise reduction over the PTI result. Fig.~\ref{fig:fusion_pipeline} shows sample results using our technique under view changes. 


Next we present a technique for relighting the free view portrait images with user specified illumination patterns. Existing single image relighting techniques ~\cite{sun2019single,pandey2021total} produce promising results but unanimously assume fixed viewpoint.
Directly applying such methods to free-view images produces inconsistent and even flicking results. We reduce such artifacts by employing the normal maps as constraints to enforce high-fidelity, consistent relighting effects across the views.

We follow a similar relighting process as ~\cite{pandey2021total} with three key modules: a normal prediction network$\networkNormal$, an albedo prediction network $\networkAlbedo$ and a relighting network $\networkRelighting$. Since \cite{pandey2021total} has not yet open sourced their solution, we therefore implemented our own version using a newly captured OLAT dataset. Specifically, we have constructed a photometric capture stage composed with 114 LED light sources synchronized with a companion 4K video camera. Using illumination multiplexing with optical flow compensation, we managed to capture dynamic OLAT video sequences at 25 fps. In our training dataset, we use 600K OLAT images covering 36 performances (18 male and 18 females) under 2810 HDR environment maps. In our experiments, we observe they are sufficient to handle a diverse class of portraits in facial shapes, complexions, skin textures (e.g., wrinkles).

We follow the ~\cite{ronneberger2015u} footprint and use U-Net as the backbones for $\networkNormal$, $\networkAlbedo$, and $\networkRelighting$ and follow the same training strategy as \cite{pandey2021total}. Since the OLAT data can be used to extract the ground truth albedo and normal maps via photometric stereo, we use the OLAT results for training the complete network. This allows us to first predict a high quality normal map $\normal$ from the input portrait $\image$, where $\normal = \networkNormal(\image)$ that has been used to support free view generation, as described in Sec.~\ref{sec:normal}. We can then apply the albedo network to extract the reflectance map $\imageAlbedo = \networkAlbedo(\image, \normal)$. Finally, we generate the relighted image $\imageRelit$ using the relighting network $\imageRelit=\networkNormal(\imageAlbedo, \normal, \lightEnv)$, where $\lightEnv$ refers to a user specified environment map.

A key different between NARRATE and TotalRelighting is that NARRATE supports relighting at a new viewpoint $\imageNote{\viewarb}$. Conceptually one can first conduct free-view rendering and then apply ~\cite{pandey2021total,sun2019single} on $\imageNote{\viewarb}$. This would require estimating a per-viewpoint normal $\normalUp{\viewarb}=\networkNormal(\imageNote{\viewarb})$ and a per-viewpoint albedo map $\imageAlbedoUp{\viewarb}=\networkAlbedo(\imageNote{\viewarb}, \normalUp{\viewarb})$. Such an implementation is not only expensive but also difficult to maintain consistency across views in normal and albedo estimation. In addition, robustly estimating normal maps from side views remains challenging, especially due to discontinuity caused by occlusions, e.g., the nose will partially occlude the cheek. 

Alternatively, similar to NARRATE, one can warp the normal field from the front view to new perspectives to void redundant computations. However, such a warping requires using 3D geometry (depth maps) where integrating the normal field relies on accurate boundary estimation. In addition, when $\viewarb$ is close to side views, warping the normal map needs to accurately resolve occlusions. In NARRATE, the front view normal map is computed using the fusion method (Sec.~\ref{sec:fusion}) whereas the boundary is provided by the SDF geometry. This allows us to conduct warping-based relighting without performing per-view normal re-estimation, i.e., $\imageRelitUp{\viewarb}=\networkNormal(\imageAlbedoUp{\viewarb}, \normalhighqualUp{\viewarb}, \lightEnv)$ where $\imageAlbedoUp{\viewarb}=\networkAlbedo(\imageNote{\viewarb}, \normalhighqualUp{\viewarb})$.

\begin{figure*}[t]
    \begin{center}
	    \includegraphics[width=0.93\linewidth]{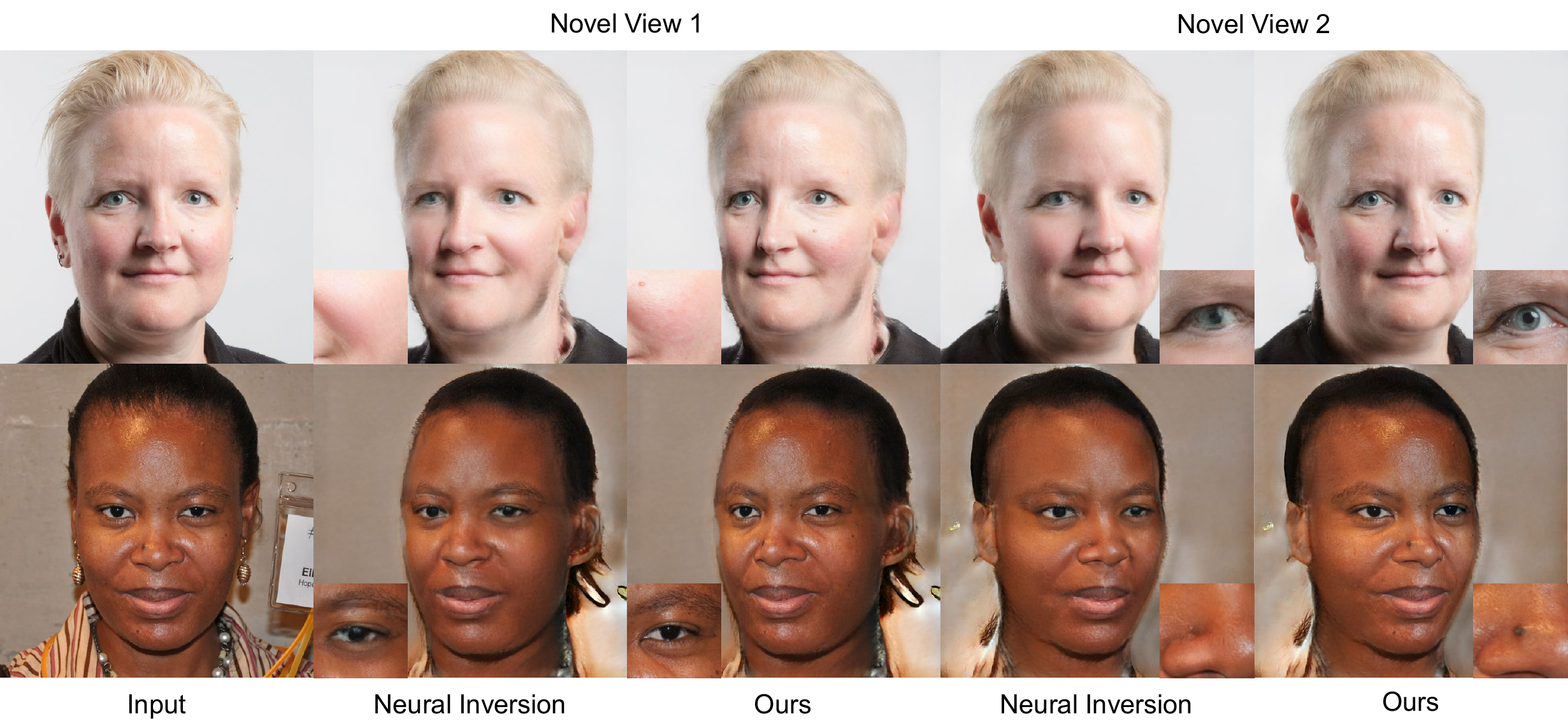}
    \end{center}
	\vspace{-0.5cm}
	\caption{\textbf{Photorealisitc Free View Synthesis Result.} Our method produces high quality portrait images of new perspectives. This method generates images with greater clarity than using the StyleSDF's Generator directly after PTI and maintains the maximum number of features of the portrait, including freckles, eyes, lips, etc.}
	\label{fig:perspective}
	\vspace{-0.3cm}
\end{figure*}



\section{Portrait Stylization}
\label{sec:application}
In addition to viewpoint changes and relighting, our hybrid neural physical facial model supports a variety of stylization. 
\subsection{Facial Animation}

{\ourmodel} can be integrated with facial animation tools for creating realistic animated avatars. 
Here, given a target video $\videoTar$ , our goal is to drive the source image $\image$ to behave the same actions and expressions as each frame in $\videoTar$ while keeping its own content identity. 

In our system, we modify the popular image animation method {\fomm}~\cite{fomm2019}. The original {\fomm} assumes that the source image and the first frame in the target video $\videoTar$ are aligned, i.e., they have to have an identical pose. This is a strong constraint that greatly limits its applicable to the types of input portrait. As {\ourmodel} supports view changes while maintaining realism, we can easy first produce a portrait to match the initial pose of the video and then conduct animation transfer. 


To reiterate, given a source image $\image$ and the $\framet$-th frame $\videoTarNote{\framet}$ in the driving video, {\fomm} detects the keypoints and estimates a backward dense motion field $\motionFieldNote{\image}{\videoTarNote{\framet}}$ in the form of per-pixel correspondence from $\videoTarNote{\framet}$ to $\image$, along with an occlusion field $\occlusionFieldNote{\image}{\videoTarNote{\framet}}$. A generator $\networkMotionGenerator$ is further employed to synthesize the animated image $\imageNote{\videoTarNote{\framet}}$, where $\imageNote{\videoTarNote{\framet}} = \networkMotionGenerator(I, \motionFieldNote{\image}{\videoTarNote{\framet}}, \occlusionFieldNote{\image}{\videoTarNote{\framet}})$. Although highly effective, the final quality using ~\cite{fomm2019} depends on accurate motion field estimations and thereby requires $\image$ and the first frame $\videoTarNote{1}$ be well aligned.

We exploit the free view generation capability of {\ourmodel} for supporting {\fomm} to drive a source image with any pose. Specifically, given the input image $\image$ and target video $\videoTar$, we first create a reference image $\imageNote{r}$ which has the same pose as $\videoTarNote{1}$. This allows {\fomm} to obtain an accurate motion field $\motionFieldNote{\imageNote{r}}{\videoTarNote{\framet}}$ between the reference image $\imageNote{r}$ and the corresponding frame $\videoTarNote{\framet}$. We then estimate the motion filed $\motionFieldNote{\image}{\imageNote{r}}$ from $\imageNote{r}$ to the source image $\image$. Recall that $\imageNote{r}$ and $\image$ are the same face rendered with different poses using the known neural-physical model. Therefore we can directly compute $\motionFieldNote{\image}{\imageNote{r}}$ from the known shape and pose changes. The complete process for motion field estimation is as follows: 
$\motionFieldNote{\image}{\videoTarNote{\framet}} = \motionFieldNote{\imageNote{r}}{\videoTarNote{\framet}} + \motionFieldNote{\image}{\imageNote{r}}$. The animated image $\imageNote{\videoTarNote{\framet}}$ is generated by $\networkMotionGenerator$ in the same way where the occlusion field is automatically computed with the motion field.

Combining {\ourmodel} and {\fomm} produces exciting new applications at an unprecedented quality: {\ourmodel} is fully automated and only requires the user provide an input portrait and the target environment for relighting; therefore we can connect our technique with any pre-trained {\fomm} models ~\footnote{https://github.com/AliaksandrSiarohin/first-order-model} to emulate a variety of facial animations, from Broadway performances to Peking Operas. All results shown in the paper are automatically generated without any fine-tuning. 

\begin{figure*}[t]
    \begin{center}
	    \includegraphics[width=0.95\linewidth]{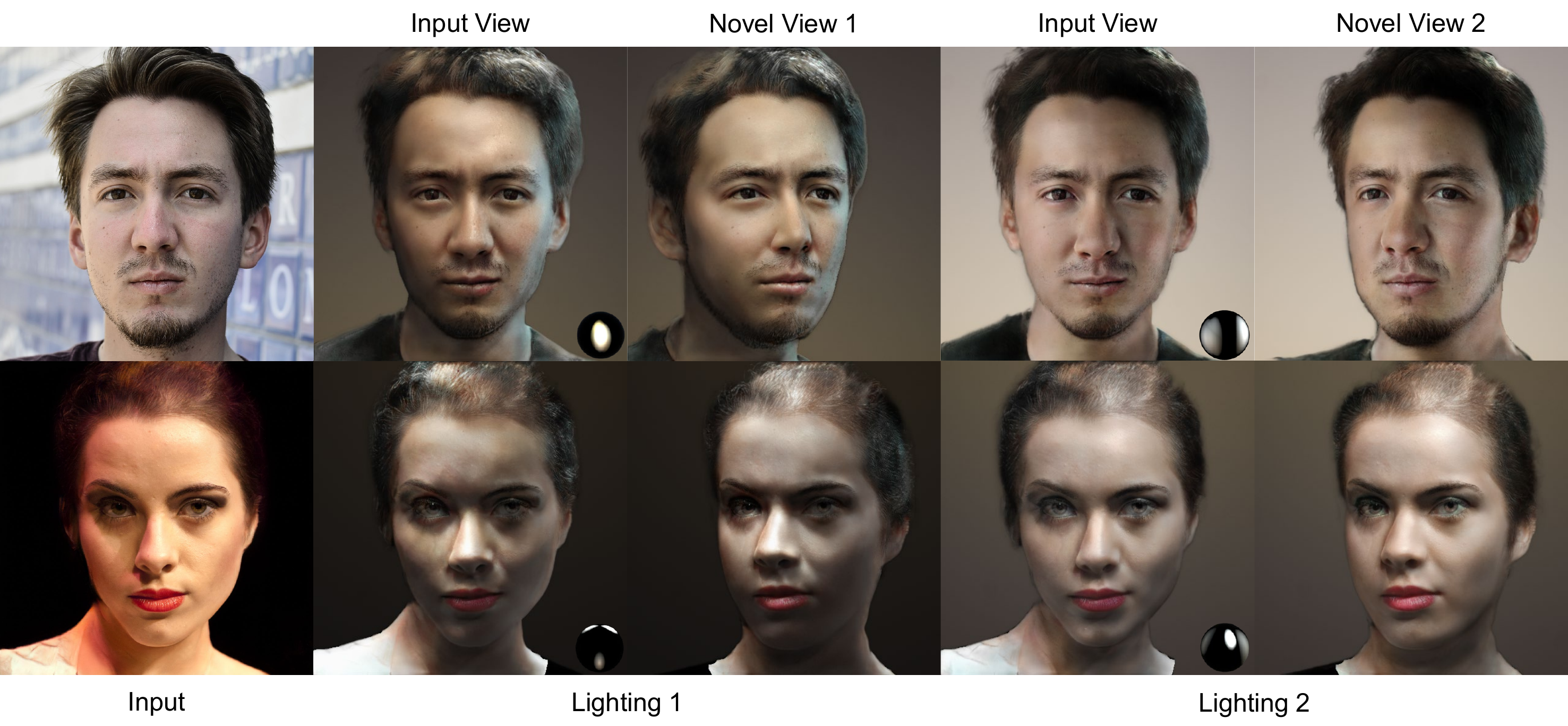}
    \end{center}
	\vspace{-0.5cm}
	\caption{
	\textbf{Consistent Free View Relighting Result.} {\ourmodel} supports consistent relighting for arbitrary views. For the input images (Col.1), we illuminate them with two environment maps. We exhibit relighted images at two views under each light (Col.2\&3, Col.4\&5). We demonstrate plausible and consistent relighting results among different views.}
	\label{fig:lighting_results}
	\vspace{-0.3cm}
\end{figure*}

Compared with state-of-the-art free view facial animation engine NTH~\cite{wang2021one}, our approach much better preserves photorealism with fine texture and geometry details. It is also faster in inference speed and more memory efficient. More importantly, we support high quality relighting that can help not only matching facial animations and poses but also lighting effects, producing dramatic mood that faithfully resembles the original clip. Fig.~\ref{fig:lighting_results} and our supplementary video show a number of examples.

\subsection{Style Transfer}

Occasionally, the pose adjusted, relighted, and even animated portrait demand a different, artistic styles. We show it is very easy to employ {\ourmodel} in style transfer, e.g., for mapping a portrait photograph to oil paintings.

State-of-the-art approaches essentially formulate style transfer as an image to image translation problem. They, however, assume aligned viewpoint and ignore lighting discrepancies. {\ourmodel}, in contrast, enables pose-aware, lighting-consistent style transfers.  

\begin{figure}[h]
    \begin{center}
	    \includegraphics[width=\linewidth]{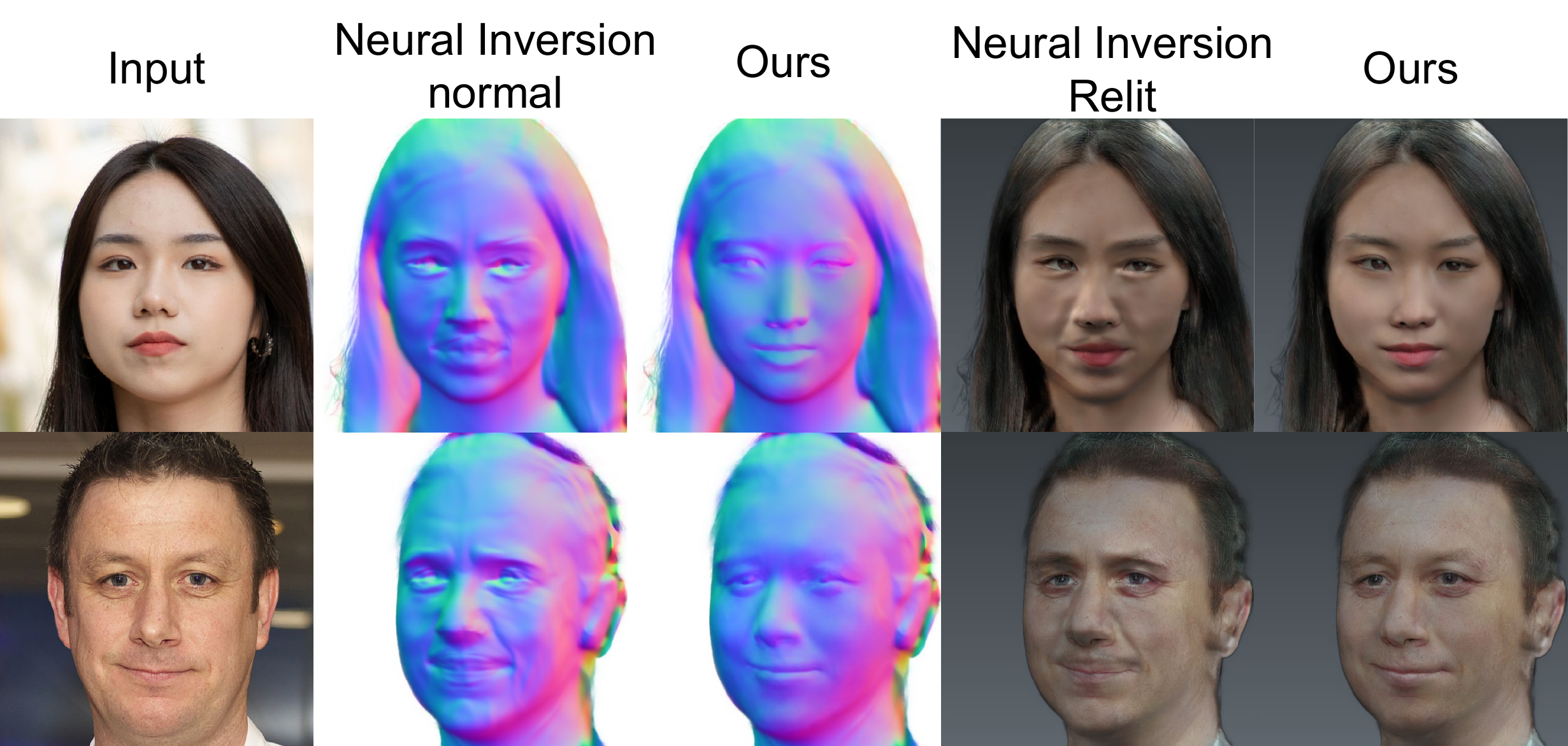}
    \end{center}
	\vspace{-0.5cm}
	\caption{
	We extract normal maps from input images with different methods
    and apply them in relighting. We compare the quality of our fused normal
    and StyleSDF normal. Clearly, our normal is more accurate and produces
    plausible relighting results, while StyleSDF extracts distorted normal and
    shows weird relighting effects.
	}
	\label{fig:relighting_comp}
	\vspace{-0.5cm}
\end{figure}

\begin{figure*}[t]
    \begin{center}
	    \includegraphics[width=0.93\linewidth]{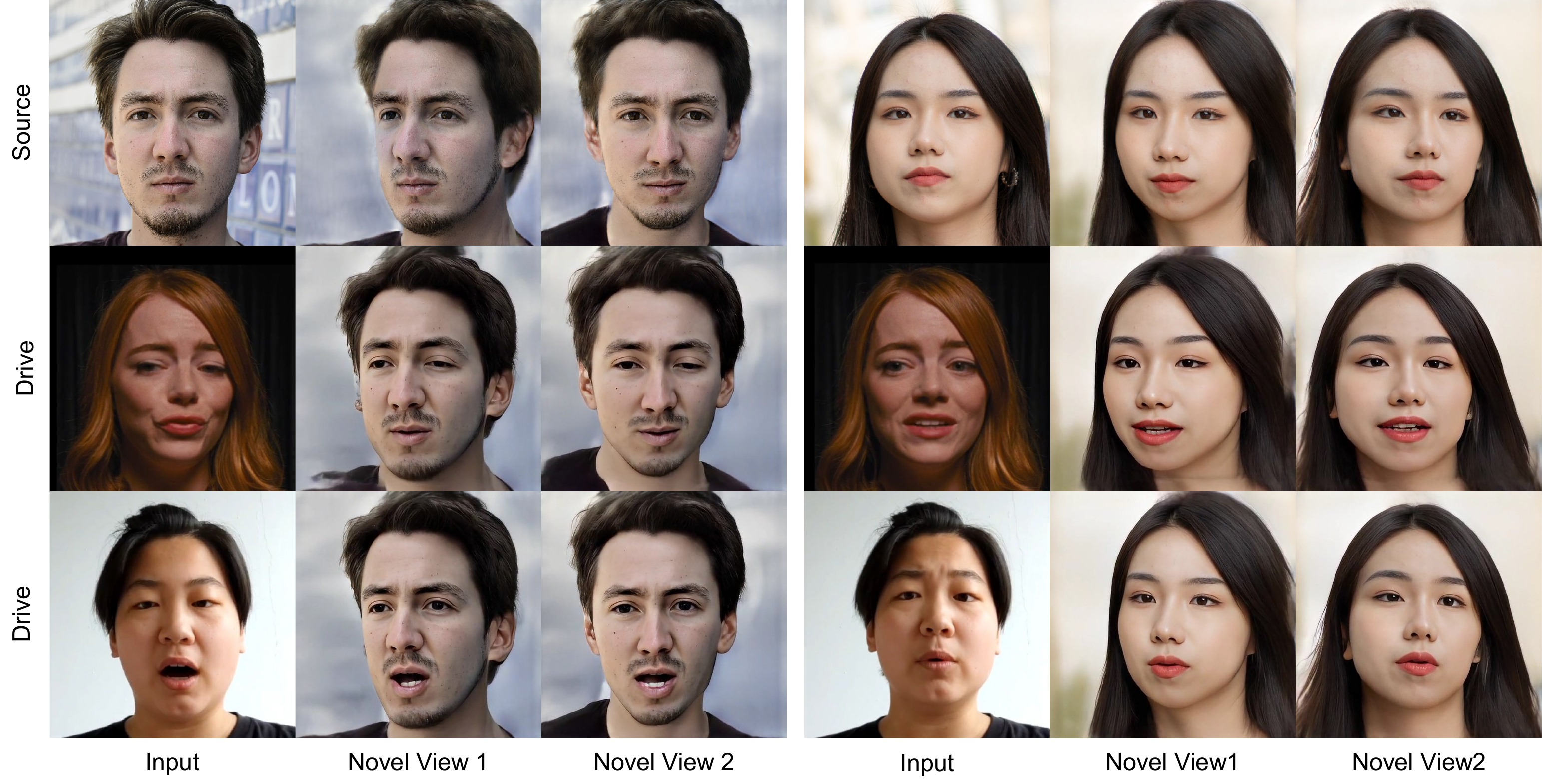}
    \end{center}
	\vspace{-0.5cm}
	\caption{
	\textbf{Free View Facial Animation.} 
		Combined with {\fomm} model, {\ourmodel} can be used to create free-view animated virtual human. Given two source images, we drive them with La La Land(Row 2) and Peking Opera(Row 3) clips. {\ourmodel} further supports synthesizing motions in novel views, creating consistent and natural free-view animation effects.
		}
	\label{fig:animation_result}
	\vspace{-0.3cm}
\end{figure*}


We follow the same warping first, transfer second procedure as in facial animation but in this case combine with a high quality image transfer engine. Given oil painting style transfer as an example, we set out train a pix2pixHD~\cite{wang2018pix2pixHD} network $\networkStyleOil$ to map a photograph $\image$ with aligned pose to oil painting style $\imageNote{\style}=\networkStyleOil(\image)$ with training data provided by ~\cite{Karras2019stylegan2}. To make sure the poses are well aligned, we again use our pose change module to render respective viewpoint that matches the one from the painting. 

What is more exciting is that the transfer network, once trained, can now support style transfers with pose changes: we simple map the portrait under the new pose to its stylized version using the same network. As {\ourmodel} maintains high view consistent, the resulting stylized sequence, although non-photorealistic, also maintains consistency. Fig.~\ref{fig:stylization} and the video shows examples of various painting styles that produce free-view portrait with corresponding styles, virtually animating an oil painting figure. 

We can also edit the lighting effects of $\imageNote{\style}$.
Notice that $\networkStyleOil$ does not readily support lighting adjustment. We therefore utilize the shading maps to implicitly relight $\imageNote{\style}$. Specifically, given a new lighting condition $\lightEnv$,  our goal is construct a relighted, stylized version $\imageNoteUp{\style}{l}$. 
We first apply our relighting module to produce a relighted image $\imageNoteUp{}{l}$ from $\image$. Next, we compute the shading map $\shading=\frac{\imageNoteUp{}{l}}{\image}$. 
Since $\imageNote{\style}$ directly comes from $\image$, applying $\lightEnv$ to $\imageNote{\style}$ can be approximated by multiplying $\imageNote{\style}$ with $\shading$ so that the final stylized result $\imageNoteUp{\style}{l} = \shading \imageNote{\style}$. Fig.~\ref{fig:stylization} also shows that our relighted oil painting portrait simultaneously preserve styles, perspectives, and lighting.

Since {\ourmodel} also provides a high quality normal map, it can be used to produce additional stylization effects that have previously relied on differential geometry operators, not easy to replicate by learning based approaches. One of such example is hatching ~\cite{praun2001real}. Real artists hatch long the two principal directions along the surface that can be computed from a high quality normal map. The tone map of the hatching texture depends on shading, which can also be adjusted by our approach. Fig.~\ref{fig:stylization} shows the sample result of a portrait stylized under hatching. 



%% file: section/Experiments.tex
\section{Experimental Results}\label{sec:results}

We evaluate our NARRATE framework on a number of portrait stylization tasks, including perspective changes, relighting, animation, style change, etc, as well as their combinations. 
We also discuss implementation details and compare them with the state-of-the-art techniques. It is worth noting that most existing approaches tackle only one specific effect whereas NARRATE can tackle multiple effects at the same time. Specifically, we show NARRATE is better at preserving find details, maintaining view consistency, producing more photorealistic relighting, on both images and videos.


\begin{figure*}[t]
    \begin{center}
	    \includegraphics[width=\linewidth]{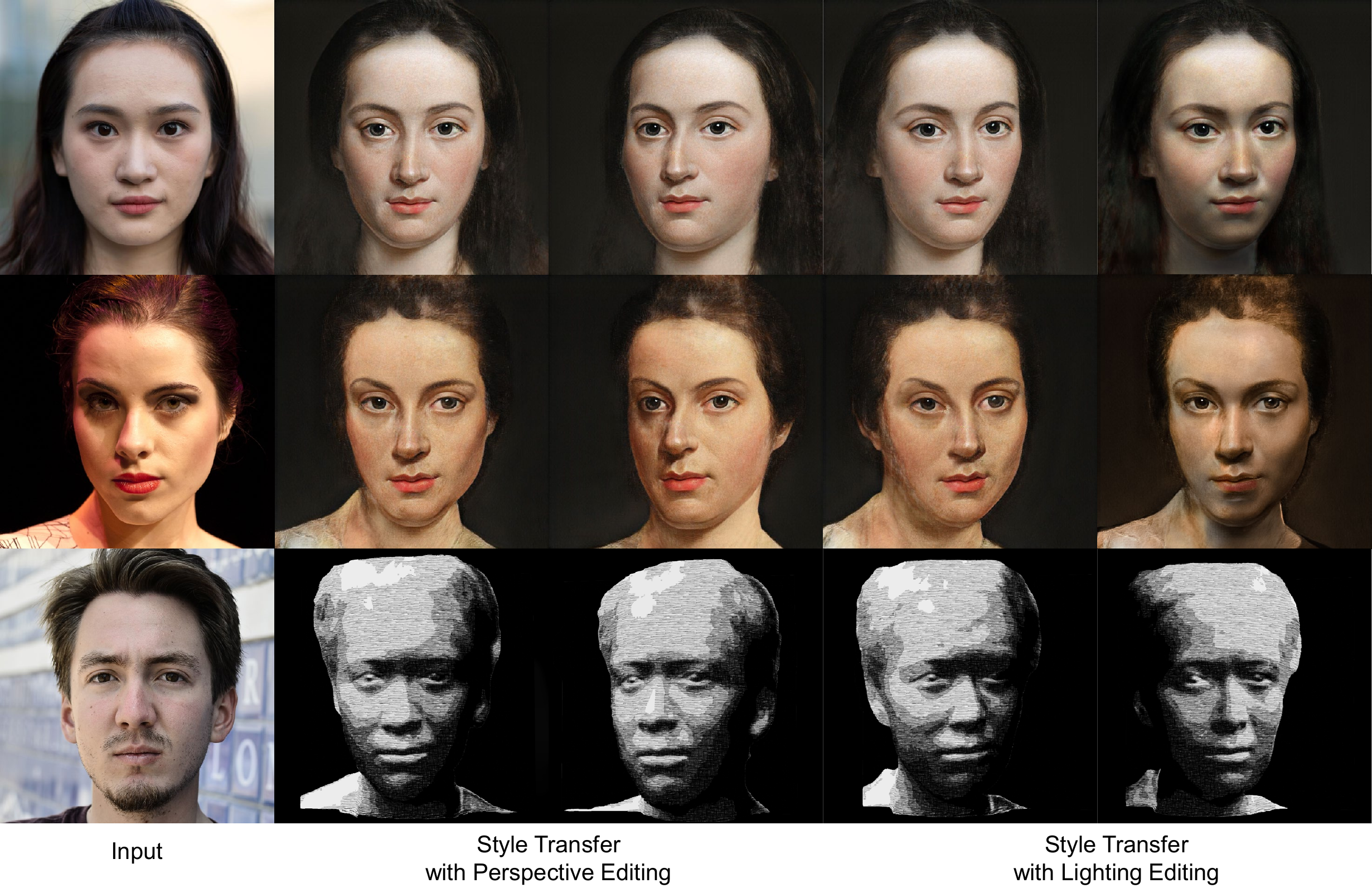}
    \end{center}
	\vspace{-0.5cm}
	\caption{
	\textbf{Style Transfer.} {\ourmodel} supports to integrate with image stylizers, producing free-view images or videos with the target style. We show oil painting in top 2 rows and hatching in the last row. We change pose in Col 2,3,4 and light in Col.5. After editing, the results still preserve the coherence of input portrait appearance.  
	}
	\label{fig:stylization}
	\vspace{-0.3cm}
\end{figure*}

\subsection{Novel View Synthesis}

We first compare NARRATE with previous approaches on perspective changes for portraits. In prior art, the same effect corresponds to novel view synthesis whereas we provide explicit pose controls. We show NARRATE, using a hybrid neural physical approach, preserves photorealism as well as coherence.  



We use the seminal work of {\stylesdf} with its pre-trained  model~\footnote{https://github.com/royorel/StyleSDF} as baseline. Recall that the original StyleSDF serves as a generator, i.e., it synthesizes images from latent codes instead of directly changing style of a given portrait. We therefore need to neural inversion. We use the latest PTI~\cite{roich2021pivotal} inversion to find the closest generated image that resembles the input portrait. To elaborate, we follow the two-stage optimization strategy: we first fix the network weights and conduct 1,000 iterations to optimize the latent code and camera view; we then fix the latent and view point and perform extra 1,500 iterations to fine-tune the network weights. 
We adopt the MSE and Perception loss between the input and the inverted images with the total loss as $\loss=\lweightMSE\lossMSE+\lweightPer\lossPer$. We use the same setting as ~\cite{roich2021pivotal} with $\lweightMSE=0.1$ and $\lweightPer=1$. Using the PTI inverted image as input to StyleSDF produces our baseline results, a free-view portrait rendering along with a coarse mesh.

Using NARRATE's physical modeling module, we infer the portrait normal as discussed in Sec.~\ref{sec:normal} and then fuse it with the coarse mesh into a high-fidelity physical face model using Poisson integration. We texture the result with the original image and use Pytorch3D~\cite{ravi2020pytorch3d} to physically render the model. Finally, we fuse the physically-rendered component (the facial part) and the neural-inverted rest (hair, ear, clothing, background, etc) into the final result.
On a desktop with an Intel Core i9-10980XE CPU (3.00 GHz) and a NVIDIA TITAN RTX GPU (24GB memory), neural inversion converges in 20 minutes  whereas the rest of the process takes only a few seconds.

Fig.~\ref{fig:perspective} compares perspective changes of three sample portrait, one for each row.  The baseline {\stylesdf}+PTI technique produces reasonable results at individual new poses but the quality still falls short of photographic quality. NARRATE, with the support of geometry-aware generation, much better preserves fine details and produces portrait-level images. For example, closeup view of the eye and mouth regions demonstrate that the baseline tends to create overly-smooth appearance with many important details washed out, which, in contrast are well preserved by our technique.  

A more severe artifact of the baseline is view consistency. In the supplementary video, we synthesize free view portrait videos both techniques. {\stylesdf} even with SDF constraints cannot fully account for detail consistency. This is largely due to the radiance field based multi-view SDF estimation is not yet capability of producing fine details, resulting in temporal (cross-view) aliasing and vibrations. In addition, such artifacts are magnified in portraits as human eyes are particularly acute to tell differences under view changes. In comparison, we impose the learned normal map, with is both high quality and free-view consistent and {\ourmodel} produces stable and aliasing-free results.

\subsection{Free View Relighting}

Next apply {\ourmodel} in free view relighting. Our approach predict plausible normal and albedo maps, achieving consistent illumination effects for arbitrary poses. 





We use the dynamic OLAT dataset to conduct end-to-end training of the three individual modules of relighting network: the albedo prediction network $\networkAlbedo$, the normal prediction network $\networkNormal$, and the relighting network $\networkRelighting$. We use the Adam optimizer~\cite{zhang2018improved} and set the learning rate to $10^{-3}$, and use SSIM with MSE loss functions. The training converges in 48 hours.

Fig.~\ref{fig:lighting_results} shows free-view relighting results on several portraits with different facial shapes and skin complexions. By applying a single normal map across different perspectives at the same time maintaining consistent 3D geometry, NARRATE achieves consistent relighting, captures facial contours, and preserves texture details.
Benefiting from the high-quality dynamic OLAT dataset, we virtually cast professional illuminations  such as Rembrandt lighting on to the subject. A unique trait of our result is that it produces photographic quality similar to \cite{pandey2021total} whereas previous single image relighting techniques tend to lose many fine details and are limited to relatively simple lighting conditions. We can further use NARRATE for image retargetting: given an input portrait and a target one, we can simultaneously match the pose and lighting. Specifically, we can first estimate the pose and illumination of the target image and then use NARRATE to transform the input portrait to the target pose with corresponding lightings. Fig.~\ref{fig:lighting_results} shows a few examples where we manage to produce photographic quality retargeting results.

\begin{table}[t]
\centering
\resizebox{\linewidth}{!}{%
\begin{tabular}{|c|c|c|c|c|}
\hline
\multicolumn{1}{|c|}{Algorithm} & \multicolumn{1}{c|}{\textless $5^\circ$} & \multicolumn{1}{c|}{\textless $15^\circ$} & \textless $25^\circ$ & \textless $30^\circ$\\ \hline
StyleSDF    & 66.637\%          & 72.069\%                  & 78.635\%    & 81.638\%       \\ \hline

\textbf{Ours}  & \textbf{67.449\%} & \textbf{75.288\%}         & \textbf{81.281\%}  & \textbf{83.5095}\% \\ \hline
\end{tabular}%
}
\caption{Normal Error on OLAT Dataset. We compare our normal and
StyleSDF normal on captured OLAT data. Our normal is more accurate
than StyleSDF.}
\label{table:normal}
\vspace{-0.5cm}
\end{table}

\begin{table}[t]
\centering
\begin{tabular}{|c|c|c|c|}
\hline
Method & SSIM$\uparrow$             & PSNR$\uparrow$               & RMSE$\downarrow$              \\ \hline
StyleSDF & 0.895                      & 23.809                     & 0.072                         \\ \hline
Ours   & \textbf{0.936}             & \textbf{26.426}              & \textbf{0.054}                \\ \hline
\end{tabular}

\caption{Quantitative comparison on portrait relighting. We then apply the
two normal prediction in relighting. No surprisingly, our better normal also
brings better relighting results.}
\label{table:com_relit}
\vspace{-0.5cm}
\end{table}

One of the major benefits of NARRATE is that it produces coherent pose changes with consistent relighting effects. As we employ a high quality normal map as constraints for relighting across viewpoints and a hybrid neural-physical face model to ensure correctness in pose-based warping, we produce very smooth and natural transitions under continuous pose changes, e.g., for emulating a face turning from the front to the side, as shown in Fig.~\ref{fig:relighting_comp} and the supplementary video. 

To demonstrate the importance of high quality normal maps, we further show the results using the normal map from {\stylesdf} for relighting. Although the overall geometry (e.g, per-view depth maps) from {\stylesdf} tends to be smooth, computing the normal maps from the estimated geometry leads to large errors, especially near eyes, cheeks, chin, etc. This is mainly because {\stylesdf} does not explicit use normal maps as constraints in either training or inference. Table.~\ref{table:normal} shows the quantitative comparisons between NARRATE and StyleSDF normals for relighting. For the ground truth, we use the OLAT data to generate 150 images under different environment illuminations. NARRATE achieves better performance in PSNR and MSE compared to using StyleSDF normal for relighting. 





\subsection{Facial Animation and Style Transfer}

The capability of free-view relighting using NARRATE can be further integrated with existing tools to support additional stylization. Specifically, we demonstrate virtual facial animation and style transfers.

\paragraph{Facial Animation.} 

Combined with {\fomm} model, {\ourmodel} can be used to create free-view animated virtual human, with broad applications in virtual and augmented reality. Given a source image $\image$ and a drive video $\videoTar$, {\ourmodel} can animate $\image$ to perform the same facial expression as $\videoTar$. In all experiments, we use the pre-trained {\fomm} model to drive the portrait without any fine-tuning. 

Fig.~\ref{fig:animation_result} shows several sample animated faces under different viewpoints. Both the La La Land and Peking Opera clips demonstrate that we can convert a static portrait to live actions, whether singing or chanting, at the photorealism. At any time instance, we can render new viewpoints of the animated performance, potentially providing an immersive viewing experience to the audience. Our current implementation cannot yet achieve near real-time performance. Yet tailored accelerations may be possible to support real-time rendering and thereby upgrading the 2D avatar animator to 3D.  

\begin{figure}[H]
    \begin{center}
	    \includegraphics[width=0.93\linewidth]{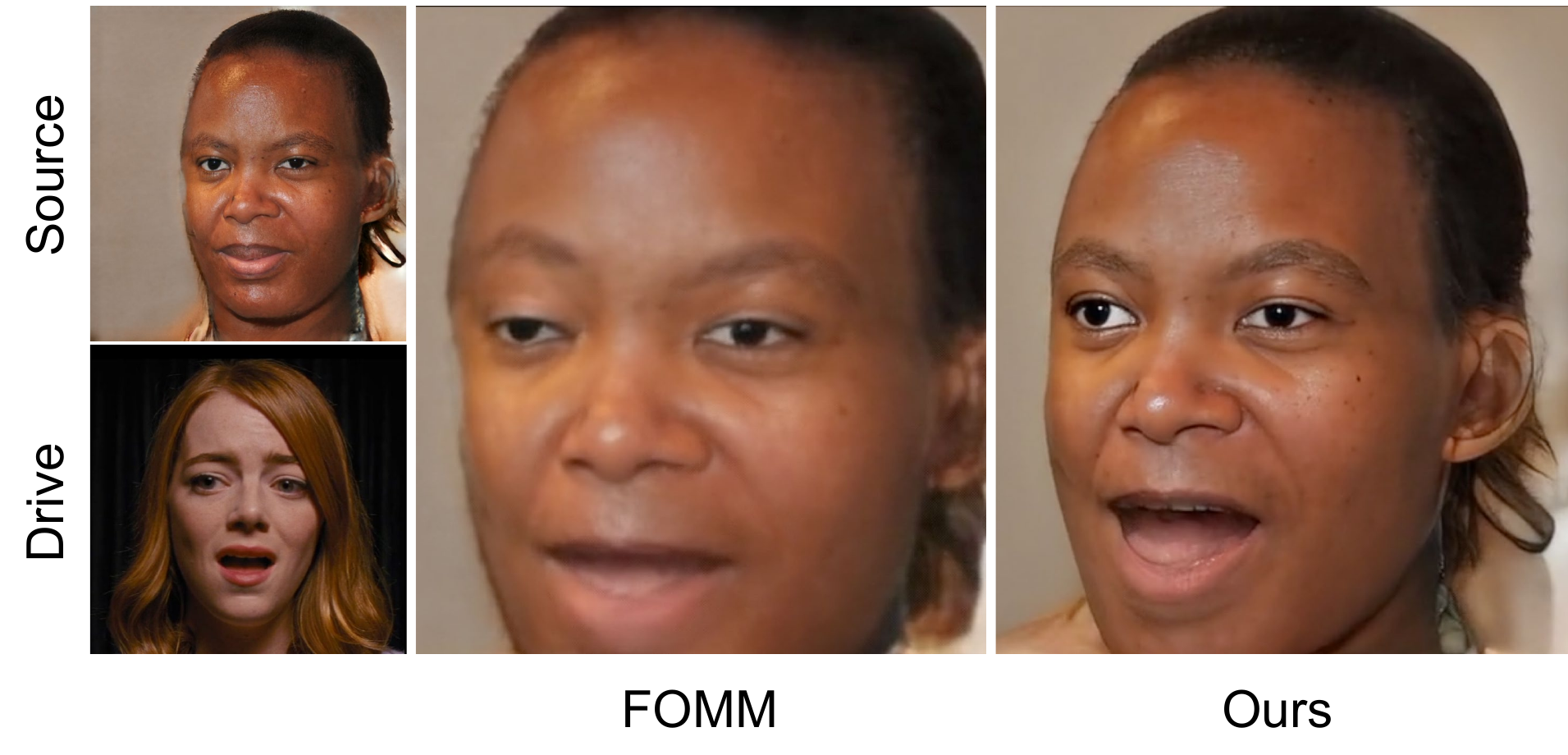}
    \end{center}
	\vspace{-0.5cm}
	\caption{
	We compared our method with {\fomm}~\cite{fomm2019}. {\ourmodel} supports large pose variations between the source and driving frames, producing animations with high-frequency details and vivid expression. On the contrary, {\fomm} fails to hand such a case, producing distorted and blurry results.
	}
	\label{fig:comp_animation}
	\vspace{-0.3cm}
\end{figure}

Another unique capability of {\ourmodel} that it lifts the restriction of the input to {\fomm}: the original {\fomm} expects the input portrait to have a pose that matches the first frame of the video $\videoTarNote{1}$. Otherwise the final generated video exhibits strong distortions due to pose discrepencies. NARRATE eliminates this requirement: we can simply render the portrait at the pose of $\videoTarNote{1}$ and use it as input to {\fomm}. Fig.~\ref{fig:comp_animation} and the supplementary video compares the animated faces with and without using NARRATE for pose correction. With pose correction, facial distortions are significantly reduced and the final results appear much more natural. 



Alternative one-shot facial animation solution such the Neural Talking Head (NHT)~\cite{wang2021one}, although effective, tend to lose facial details. This is mainly due to they do not employ full 3D facial geometry in the synthesis process. For example, NTH adopts a keypoint representation to decompose identity-specific and motion-related information and it can achieve robust pose synthesis under rotation as well as complex facial animation. However, keypoints alone are insufficient to maintain detailed geometry or appearance and their results cannot meet the quality of high resolution videos, e.g., for feature film productions. In contrast, by inferring a high quality neural physical face model, NARRATE can faithfully preserve details critical to preserve facial traits. NARRATE further provides effective relighting capabilities absent in prior art, to produce unprecedented lighting effects at a photographic quality. Fig.~\ref{fig:animation_result} and our supplementary video show a number of examples.

\paragraph{Style Transfer.}

{\ourmodel} maintains coherence of appearance under view and lighting changes as well as with animations. When combined with an image stylizer, it produces free-view point images or videos with the target style. Three are a number of high resolution image stylization tools based on either on classical physically-based rendering or emerging generative models. The former uses tailored shaders for processing known geometry primitives, e.g., hatching requires smooth surfaces to robustly estimate the hatching directions. The latter such as PixeltoPixelHD ~\cite{wang2018pix2pixHD} takes images as inputs without known geometry and can produce a wide variety of styles. As a hybrid neural physical model, {\ourmodel} supports both types of stylization. 

Fig.~\ref{fig:stylization} and the supplementary video show the typical examples of synthesizing the oil painting painting and hatching. For oil painting, we use {\ourmodel} to produce a single image under the original or new lighting or a video sequence with continuous pose changes with fixed lighting or even track light with the movement. We then feed the results into our syylizer, the pix2pixHD network, to stylize individual frames. Combined with facial animation, style transfers on top of {\ourmodel}'s free-view rendering can bring famous oil painting figures to live and viewable in 3D, providing surreal experiences to users. 

Since most styles serve as de facto low-pass filters, they preserve perspective and temporal coherence of {\ourmodel} generated results. A notable exception of hatching that produces additional high frequency details rather than smoothing. Applying image to image style transfer to emulate hatching effect cannot reach the quality as in classic rendering based methods, e.g., hatching should align with the principal directions on the mesh. Since {\ourmodel} produces a high quality normal map, it is straightforward to directly compute the hatching directions to produce visually compelling results. More importantly, since the normal map is coherent across view direction changes, conducting geometry-based hatching on free-view portraits using {\ourmodel} preserves view consistency, avoiding disturbing shower-door effect common in generation based results.

Finally {\ourmodel} can unify pose change, relighting, facial animation, and style transfer into a fully automated pipeline, achieving these effects either separately or in combination, all at a photographic quality. Fig.~\ref{fig:stylization} shows an example with all four effects: a single portrait, we produce an oil painting style free-view video of the subject performing opera. In an Andy Wahol style matrix layout, we vary pose in one dimension and lighting the other. Each synthesized image maintains fine details in facial traits and hence identity and at the same time transitions across the matrix appear coherent and harmonious. Additional results can be found in the supplementary video.





\begin{figure}[t]
    \begin{center}
	    \includegraphics[width=0.93\linewidth]{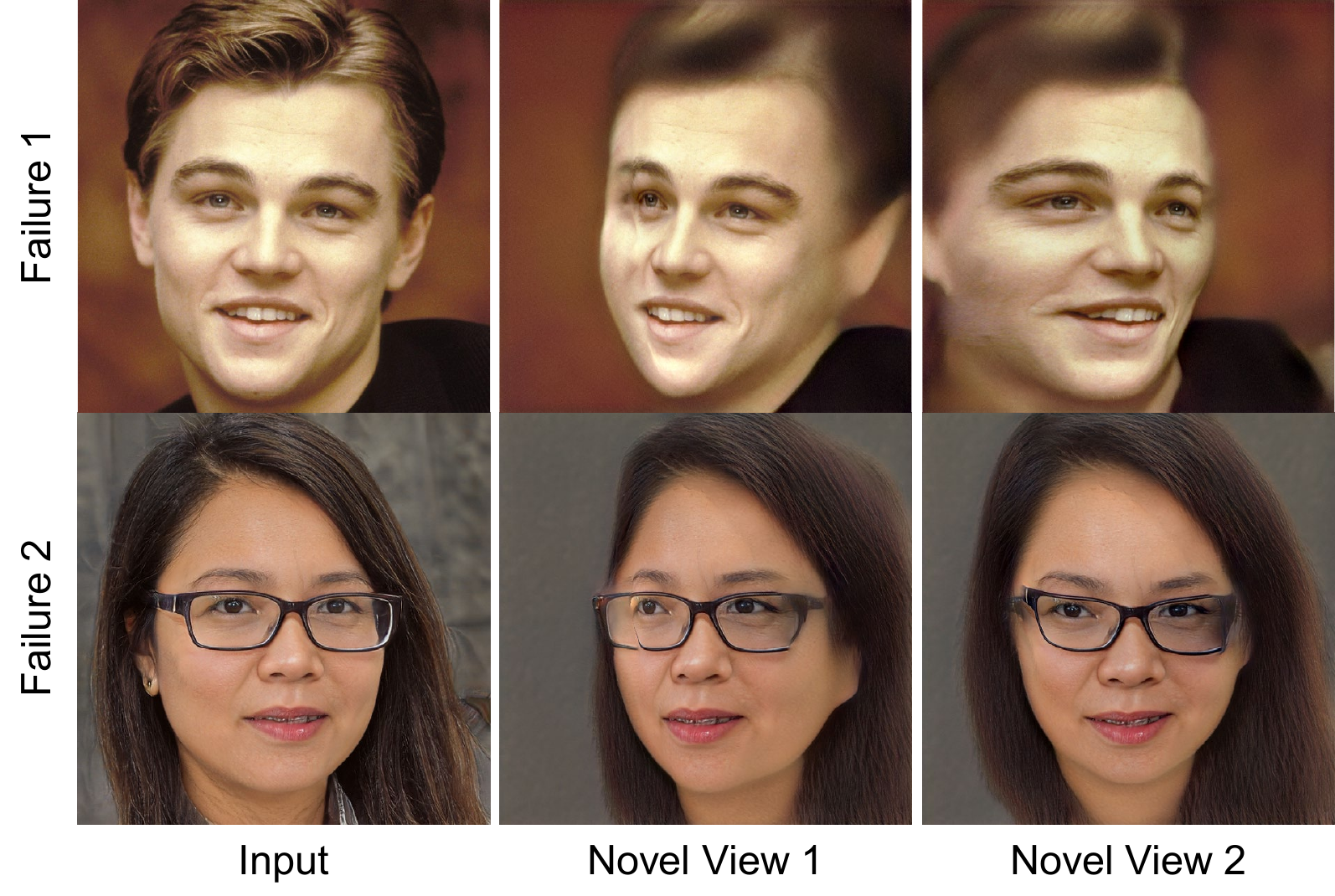}
    \end{center}
	\vspace{-0.5cm}
	\caption{
	We show two failure cases. 
		The first one contains odd camera angle that is beyond the training distribution. As a result, our neural model fails to run invert it to a reasonable view and geometry. 
		The glasses regions in the second example presents us from exacting correct normal for our physical model, also resulting in a distortion when changing views.
		}
	\label{fig:failure_case}
	\vspace{-0.3cm}
\end{figure}

\subsection{Limitations}



We have verified {\ourmodel} on a variety of applications, providing a new scheme for high quality portrait editing. However, as the first pipeline to accomplish all the goals, {\ourmodel} still has the following limitations, as shown in Fig.~\ref{fig:failure_case}.
Firstly, our method relies on the inherent capabilities of the geometry-aware generative models, which would fail in the cases beyond the training distributions. E.g., when the camera angle is too lateral, {\stylesdf} produces poor fitting, leading the failure of {\ourmodel} as well. 
In addition, our physical model is built on top of the coarse geometry obtained from {\stylesdf}. However, the correctness of geometry is also limited in corner cases, especially when the person is wearing adornments (e.g. glasses, hat). This also influences accuracy of the physical model, bringing incorrect normal estimation and introducing artifacts when conducting photo editing. This problem existed in both neural part and physical part would be a valuable research topic in our future work.

\begin{figure*}[ht]
    \begin{center}
	    \includegraphics[width=0.95\linewidth]{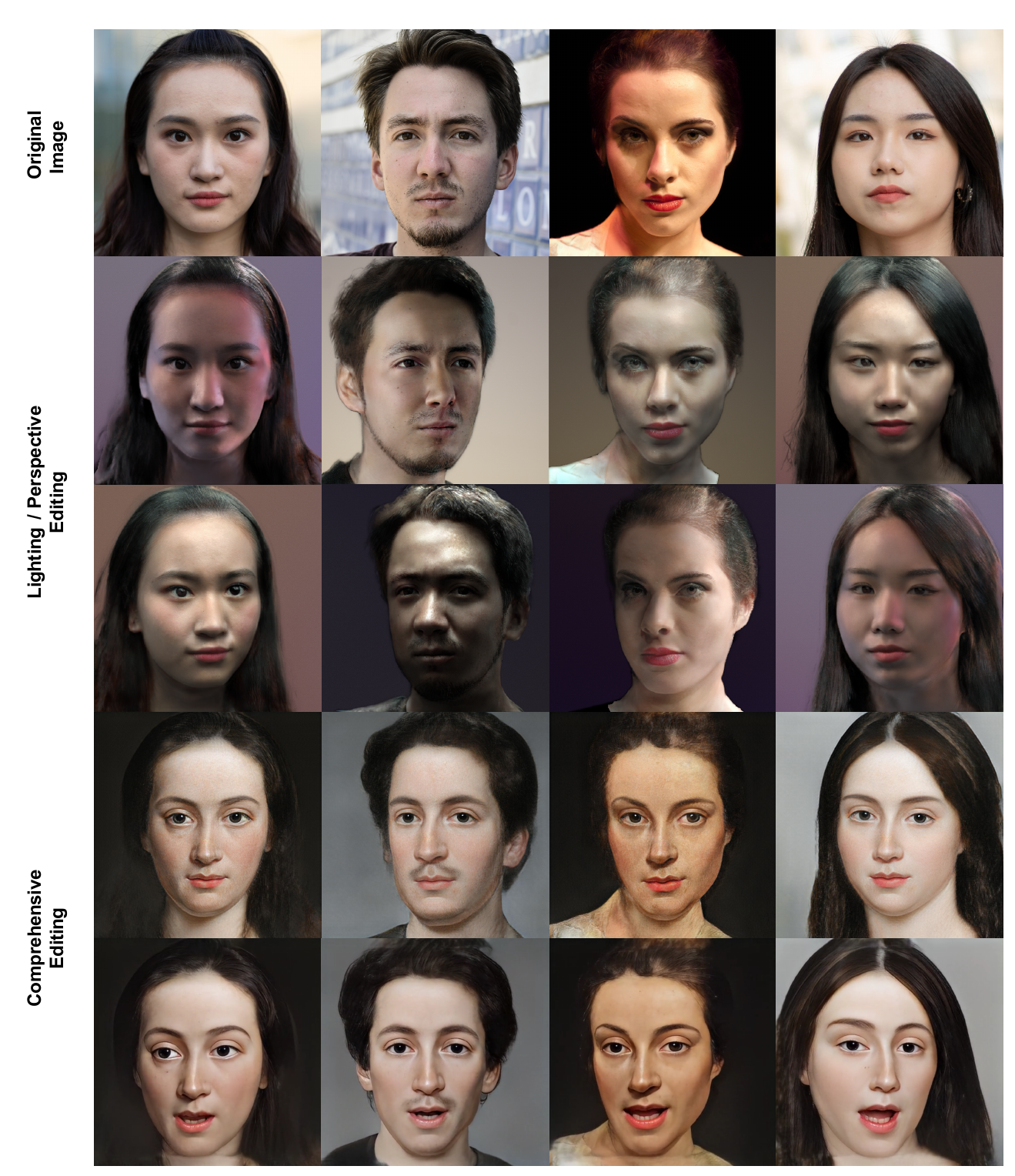}
    \end{center}
	\vspace{-0.5cm}
	\caption{
	\textbf{Comprehensive editing.} We demonstrate the comprehensive editing ability of {\ourmodel}. It can unify pose change, relighting, facial animation, and style transfer into a fully automated pipeline, achieving these effects either separately or in combination, all at a photographic quality. In row 2\&3, we show free-view relighting. In row 4\&5, we combine all the editing effects together.}
	\label{fig:Gallery}
	\vspace{-0.3cm}
\end{figure*}